\PassOptionsToPackage{table,usenames,dvipsnames}{xcolor}
\documentclass{article}
\usepackage{ijcai25}
\usepackage{pifont}

\usepackage{bbm}
\usepackage{multirow}
\usepackage{times}
\usepackage{soul}
\usepackage{url}
\usepackage{subcaption}
\usepackage[hidelinks]{hyperref}
\usepackage[utf8]{inputenc}
\usepackage{caption}
\usepackage{graphicx}
\usepackage{xcolor}
\usepackage{amsmath}
\usepackage{amsthm}
\usepackage{booktabs}
\usepackage{algorithm}
\usepackage{algorithmic}
\usepackage{natbib} 
\usepackage[switch]{lineno}
\usepackage{amsmath}
\usepackage{amssymb}
\usepackage{rotating}
\usepackage{enumitem} 
\setlist[itemize]{leftmargin=7pt,labelsep=3pt,noitemsep,topsep=5pt}
\setlist[enumerate]{leftmargin=20pt,labelsep=10pt,noitemsep,topsep=5pt}

\title{Knowledge-Informed Deep Learning for Irrigation Type Mapping from Remote Sensing}

\author{
Oishee Bintey Hoque\textsuperscript{\rm 1,\rm *} \and
Nibir Chandra Mandal\textsuperscript{\rm 1,\rm *} \and
Abhijin Adiga$^{2}$ \and
Samarth Swarup$^2$ \and
Sayjro Kossi Nouwakpo$^3$ \and
Amanda Wilson$^2$ \and
Madhav Marathe$^{1,2}$ \\
\affiliations
$^1$Dept. of Computer Science, University of Virginia \\
$^2$Biocomplexity Institute, University of Virginia \\
$^3$US Department of Agriculture \\
\emails
\{gza5dr, wyr6fx, abhijin, swarup, alw4ey, marathe\}@virginia.edu,\\
Kossi.Nouwakpo@usda.gov
}

\begin{document}

\maketitle

\begin{abstract}
Accurate mapping of irrigation methods is crucial for sustainable agricultural 
practices and food systems. However, existing models that rely solely on spectral
features from satellite imagery are ineffective due to the complexity of agricultural 
landscapes and limited training data, making this a challenging problem. We present 
Knowledge-Informed Irrigation Mapping (KIIM), a novel Swin-Transformer based approach that uses $(i)$ a specialized projection matrix to encode crop to 
irrigation probability, $(ii)$ a spatial attention map to identify agricultural lands 
from non-agricultural lands, $(iii)$ bi-directional cross-attention to focus complementary information from different modalities, and $(iv)$ a weighted ensemble for 
combining predictions from images and crop information. Our experimentation on five
states in the US shows up to 22.9\% (IoU) improvement over baseline with a 71.4\%~(IoU) 
improvement for hard-to-classify drip irrigation. In addition, we propose a two-phase transfer learning approach to enhance cross-state irrigation mapping, achieving a 51\% IoU boost in a state with limited labeled data. The ability to achieve baseline performance with only 40\% of the training data highlights its efficiency, reducing the dependency on extensive manual labeling efforts and making large-scale, automated irrigation mapping more feasible and cost-effective. Code: https://github.com/Nibir088/KIIM 
\footnotetext[1]{*Both authors contributed equally to this research.}

\end{abstract}

\section{Introduction}

\textbf{Mapping Irrigation Assets: A Social Good Problem.} 
Irrigation is a crucial component of agricultural management, supporting approximately 40\% of global food production \citep{WWAP2019}. As a dominant freshwater-use practice, irrigation accounts for nearly 90\% of global consumptive freshwater use \citep{Doll2009,Meier2018,Dieter2018,Zhou2020}, significantly shaping regional and global hydrological cycles \citep{de,leng2014modeling}. In regions like northwestern China and the US High Plains, excessive irrigation has caused substantial declines in river discharge and groundwater levels, highlighting the impact of inefficiency of irrigation methods \citep{hao2015quantitative,perez2020irrigation}. The irrigation type (e.g., drip, sprinkler, or flood irrigation) determines how extracted water is distributed across irrigated areas, affecting water quantity and quality \citep{Ippolito}. While accurate mapping of irrigation methods can facilitate identification of current practices and sustainable upgrades, traditional approaches only distinguish irrigated from non-irrigated fields or focus only on small areas or a single type of irrigation; large-scale cross-region generalization remains underexplored \citep{tang2021mapping, Nouwakpo, hoque2024irrnet}. Thus, effective irrigation mapping aligns with the United Nations Sustainable Development Goals 2 and 8 \citep{UN2015}, which aim to promote sustainable agricultural practices and food systems, while also supporting the goal of ``Leave No One Behind" \citep{WWAP2019}.

\noindent\textbf{Team.}
This work is an interdisciplinary collaboration between computer scientists (at universities) and an agricultural scientist (at 
the US Department of Agriculture).

\begin{figure}[t]
    \centering
        \centering
        \includegraphics[width=9cm, height=3cm]{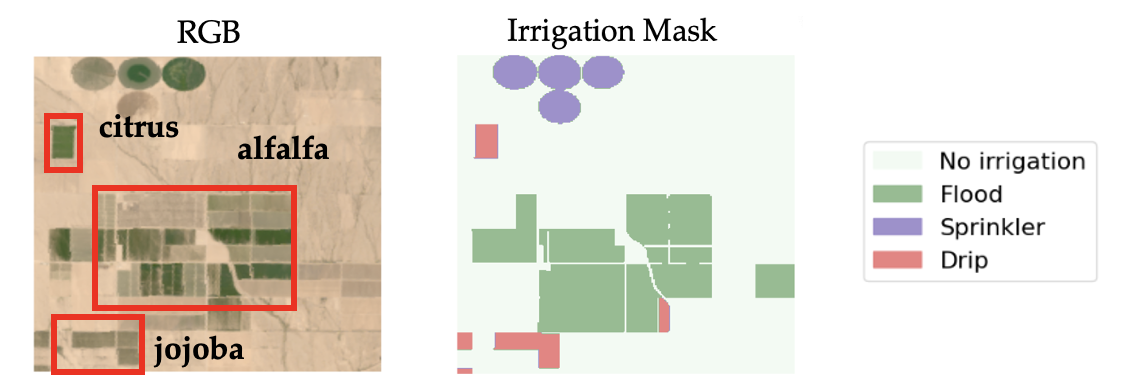}
    \caption{Irrigation mapping from satellite imagery: (left) RGB image showing citrus, alfalfa, and jojoba fields (red borders); (right) irrigation classification mask with Sprinkler, Flood, and Drip irrigation methods.}
    \label{fig:three-graphs}
\end{figure}



\noindent\textbf{Challenges.}
Due to differences in climate, crop types, and water availability, irrigation practices can vary significantly from one region to another \citep{Nie2020}. As a result, traditional methods that perform well locally may struggle to accurately classify irrigation across multiple regions, making large-scale, consistent irrigation mapping an ongoing challenge. Many regions lack sufficient labeled data for training robust irrigation classification models, as collecting ground truth across vast farmland areas is costly and time-consuming. Remote sensing offers a scalable solution using multispectral satellite imagery, but challenges persist due to spectral ambiguity, resolution limits, class imbalance, and regional differences. Furthermore, available data are often highly imbalanced, with drip-irrigated fields constituting only a small fraction of samples, making it difficult for the model to learn minority classes effectively. For instance, drip irrigation accounts for just \textbf{0.08\%} of Utah's irrigated land\footnote{\url{https://dwre-utahdnr.opendata.arcgis.com/pages/wrlu-data}}.

\noindent
\textbf{Our Contributions.}
To address these challenges, we propose the \textbf{K}nowledge-\textbf{I}nformed \textbf{I}rrigation \textbf{M}apping (KIIM) model (Fig. \ref{fig:proposed-methods}), which leverages the Normalized Difference Vegetation Index (NDVI), the Normalized Difference Water Index (NDWI), and the Normalized Difference Tillage Index (NDTI) derived from additional remote sensing bands (details in Appendix) to capture plant health, water content, and soil conditions. KIIM also uses land-use data and crop-type information, from the USDA Cropland Data Layer, to refine predictions by focusing on cultivated areas and incorporating crop-irrigation relationships, improving the identification of underrepresented irrigation methods like drip irrigation. Our main contributions are:

\begin{itemize}    
    \item  We propose a vision transformer-based multi-stream learning framework that integrates RGB and vegetation indices using a \emph{Bidirectional Cross-Attention} module (Fig. \ref{fig:proposed-methods}). Vegetation indices are closely related to irrigation methods; for example, NDWI identifies flood-irrigated fields better, while NDVI distinguishes healthy crop growth associated with sprinkler or drip irrigation \citep{geographyrealm_ndwi,utah2021drip}. While straightforward single-stream channel stacking may lead to the loss of modality-specific information due to early fusion, our framework instead guides the model to capture complementary relationships allowing each stream to query relevant information from the other stream's perspective.
\item We encode crop-irrigation relationships in a state-specific projection matrix, capturing historical irrigation preferences, and use weighted ensemble method with the prediction from the multi-stream module. Crop-type information further refines predictions, as certain crops are historically associated with specific irrigation methods (e.g., vineyards with drip irrigation, alfalfa with flood irrigation; see Fig \ref{fig:three-graphs}).
\item We incorporate a spatial attention map to enhance agricultural land segmentation by generating pixel-level attention map. This assigns higher weights to agricultural regions and field boundaries while suppressing non-agricultural areas, effectively highlighting irrigation-relevant features.
 \item Our extensive evaluations across five states (Arizona (AZ), Colorado (CO), Utah (UT), Washington (WA), and Florida (FL)) show that each module in KIIM improves performance individually, with the best results achieved when all modules are combined. KIIM consistently outperforms the baseline, achieving an average IoU improvement of 18.1\% across states and a 71.4\% improvement in challenging cases (i.e., drip). KIIM demonstrates strong generalization to unseen state data, achieving impressive performance in both zero-shot and few-shot settings. Impressively, KIIM achieves this performance using Landsat’s 30m resolution, demonstrating its ability to learn irrigation patterns despite coarse spatial granularity.
 \end{itemize}

\section{Related Work}
\noindent\textbf{Remote Sensing for Mapping Agricultural Infrastructure.}  
Deep learning models in agricultural remote sensing have been applied to crop classification, field boundary detection, and irrigation mapping \citep{jin2017mapping, weiss2020remote,paolini2022classification}. CNNs have been used for classifying invasive species \citep{hung2014feature}, segmenting mixed crops \citep{mortensen2016semantic}, and detecting weeds \citep{milioto2017real,di2017automatic}. Attention-based models \citep{wang2020deep,zheng2021rethinking} and channel-wise feature selection techniques \citep{cheng2021per,tao2020hierarchical} improve segmentation accuracy in complex landscapes. However, CNNs and ViTs struggle to differentiate irrigation methods due to spectral similarity. Multi-stream fusion of spectral indices (NDVI, NDWI, NDTI) enhances segmentation robustness but still demands large labeled datasets for reliable generalization \citep{he2017mask,touvron2021training}. While progress has been made in other agricultural infrastructure, advancements in irrigation mapping remain limited.

\noindent\textbf{Multi-Channel Representation Learning.} Recent advances in deep learning have significantly improved semantic segmentation, evolving from FCNs to encoder-decoder models like U-Net, LinkNet, and DeepLabv3+ \citep{long2015fully, ronneberger2015u, chen2017deeplabv3, chaurasia2017linknet}. Multi-scale feature extraction is enhanced by FPN and PAN \citep{li2018pan}, while transformer-based models (ViTs, Swin-Transformers) improve global context learning \citep{dosovitskiy2021image, he2022swin}. Attention mechanisms, such as CBAM \citep{woo2018cbam}, refine feature representation, and multi-stream architectures leverage attention for modality fusion in remote sensing \citep{wang2020eca, Bastidas_2019_CVPR_Workshops}. Unlike prior works that stack agricultural indices with RGB \citep{hoque2024irrnet,Nouwakpo}, our approach employs bidirectional cross-attention within a multi-stream architecture to enable dynamic feature interaction 
between streams.

\noindent\textbf{Domain-Aware Segmentation Models.}  
Remote sensing in agriculture faces domain shifts due to variations in soil, cropping patterns, irrigation, and climate \citep{raei2022deep, wang2024cross, chen2024change}. Transfer learning techniques, including feature adaptation and domain-aware fine-tuning, help align feature representations across regions 
\citep{zhuang2020comprehensive, bosilj2020transfer, coulibaly2019deep}. Recent studies show that incorporating domain knowledge enhances generalization and reduces dependence on large labeled datasets \citep{shi2021domain}.

\section{Knowledge-Informed Irrigation Mapping}
\begin{figure*}[!ht]
    \centering
    \includegraphics[width=0.838\linewidth]{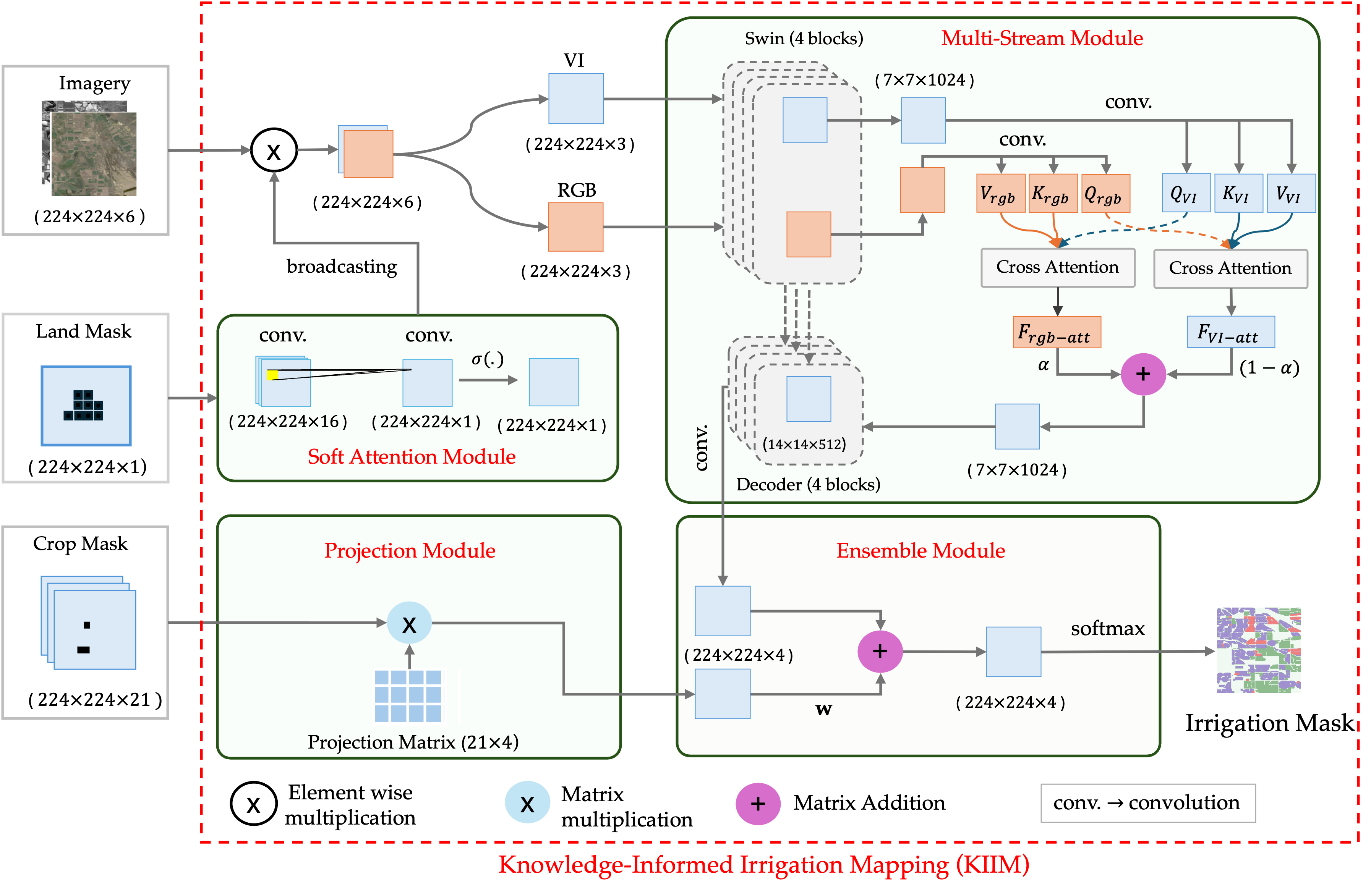}
    \caption{ Overview of our Knowledge-Informed Irrigation Mapping model. (i) A soft attention module refines spatial focus by highlighting irrigated areas. (ii) The multi-stream module processes RGB and vegetation indices through a shared Swin Transformer and fuses features using bidirectional cross-attention, allowing adaptive feature interaction. (iii) A projection module incorporates domain knowledge by mapping crop types to irrigation probabilities using a predefined projection matrix. (iv) Finally, an ensemble module balances satellite-derived predictions with knowledge-informed irrigation likelihoods, optimizing weights through end-to-end training.
    }
    \label{fig:proposed-methods}
\end{figure*}
\subsection{Problem Formulation}

We formulate irrigation mapping from satellite imagery as a semantic segmentation problem. Given a satellite image \( X \in \mathbb{R}^{H \times W \times C} \), crop mask \( M \in \{0,1\}^{H\times W \times G} \), and land mask \( L \in \{0,1\}^{H\times W} \), where \( H, W, C, \) and \( G \) denote height, width, spectral bands, and crop groups, respectively, our goal is to classify each pixel into one of \( K \) irrigation methods.

Let \( Y \in \mathcal{Y}^{H \times W} \) be the ground truth irrigation labels, where \( Y_{i,j} \) represents an irrigation type for pixel \((i,j)\). We aim to learn a mapping function:
\[
f_{\theta}: \mathbb{R}^{H \times W \times C} \times \{0,1\}^{H\times W\times G} \times \{0,1\}^{H \times W} \rightarrow [0,1]^{H \times W \times K}
\]

parameterized by \( \theta \), that outputs a probability distribution over irrigation methods. The optimal parameters \( \theta^* \) are obtained by minimizing a loss function \( \mathcal{L} \):

\begin{equation}
\theta^* = \arg \min_{\theta} \mathcal{L}\big(f_{\theta}(X,M,L), Y\big)
\end{equation}




\subsection{Methodology}

Our knowledge-informed irrigation mapping (KIIM) model addresses the key challenges in irrigation mapping through a specialized architecture that combines satellite imagery with crop knowledge. The model takes three inputs to capture different aspects of agricultural landscapes: $(i)$ Land mask (224×224×1), a binary mask defining agricultural boundaries; $(ii)$ Satellite bands (224×224×B), multi-spectral imagery providing spectral reflectance across B wavelength bands; and $(iii)$ Crop mask (224×224×21), a one-hot encoded spatial representation where each channel corresponds to a specific crop type. These inputs are processed through specialized modules designed to maximize irrigation mapping accuracy. Our architecture comprises four main modules: $(i)$ soft attention module, $(ii)$ multi-stream module, $(iii)$ projection module,  and ($iv$) ensemble module. 
Figure~\ref{fig:proposed-methods} illustrates the complete architecture.
\begin{itemize}
    \item \textbf{Soft Attention Module} applies sequential $3\times3$ convolutions followed by sigmoid activation on the land mask to generate a spatial attention map that highlights irrigated areas. In contrast to hard attention, it preserves spatial continuity, refining boundaries and preventing abrupt transitions between irrigated and non-irrigated areas. By adaptively scaling pixel-wise attention weights, it filters out background noise while increasing the contrast between irrigated and non-irrigated regions.
    \item \textbf{Multi-Stream Module} processes two complementary input streams through a shared Swin Transformer \( \phi \): RGB data (\( X_{\text{RGB}} \in \mathbb{R}^{H \times W \times 3} \)) and derived vegetation indices (NDVI, NDTI, and NDWI) (\( X_{\text{VI}} \in \mathbb{R}^{H \times W \times 3} \)). The extracted feature maps \(F_{\text{RGB}}, F_{\text{VI}}\) preserve modality-specific features while enhancing fusion and reducing overfitting compared to a single-stream model. Instead of na\"{i}ve concatenation or averaging for merging these feature maps before it goes to the decoder, we introduce \emph{\textbf{bidirectional cross-attention}}, allowing RGB and vegetation indices to exchange information and dynamically prioritize relevant features across streams, leading to more context-aware feature fusion. Unlike spatial and self-attention, bidirectional cross-attention enables explicit feature interaction between RGB and spectral streams, allowing each stream to dynamically query and integrate complementary information from the other stream~\citep{tan2019lxmert}, rather than just focusing on spatial relationships within individual feature maps or single feature spaces. To achieve this, we first transform RGB and VI features into queries (\(Q\)), keys (\(K\)), and values (\(V\)) using \(1 \times 1\) convolutions for both streams. The RGB stream is projected into \(Q_{\text{rgb}}, K_{\text{rgb}}, V_{\text{rgb}}\), while the VI stream is mapped to \(Q_{\text{VI}}, K_{\text{VI}}, V_{\text{VI}}\) with analogous definitions for $Q_{\mathrm{aux}},\,K_{\mathrm{rgb}},\,V_{\mathrm{rgb}}$. 
 Next, we calculate the standard scaled dot-product attention operation as done by~\cite{vaswani2017attention}.
For brevity, consider a single query--key--value triplet 
$(\mathbf{Q}, \mathbf{K}, \mathbf{V}) \in \mathbb{R}^{B \times C \times H \times W}$
to compute $\mathrm{Attention}(\mathbf{Q}, \mathbf{K}, \mathbf{V}) = 
\mathrm{softmax} \bigl(\tfrac{\mathbf{Q} \mathbf{K}^\top}{\sqrt{C}}\bigr) \mathbf{V}$. 
In our two-stream setting, RGB attends to vegetation indices as 
\textit{$\mathbf{F}_{\mathrm{VI\text{-}att}} = \mathrm{Attention}(\mathbf{Q}_{\mathrm{rgb}}, \mathbf{K}_{\mathrm{VI}}, \mathbf{V}_{\mathrm{VI}})$}, 
while vegetation indices attend to RGB as 
\textit{$\mathbf{F}_{\mathrm{rgb\text{-}att}} = \mathrm{Attention}(\mathbf{Q}_{\mathrm{VI}}, \mathbf{K}_{\mathrm{rgb}}, \mathbf{V}_{\mathrm{rgb}})$}. 
We then combine these attended feature maps with a learnable fusion parameter $\alpha \in \mathbb{R}$ (initialized to $0.8$), 
resulting in the final fused representation $\mathbf{F}_{\mathrm{fused}} = \alpha\,\mathbf{F}_{\mathrm{rgb\text{-}att}} + (1 - \alpha)\,\mathbf{F}_{\mathrm{VI\text{-}att}}$. Lastly, we employ a U-Net-style decoder~\citep{ronneberger2015u} with skip connections to reconstruct pixel-level predictions. A final \(1\times1\) convolution generates logits for segmentation into four irrigation classes.

    \item \textbf{Projection Module} maps domain knowledge (i.e., crop information) to irrigation probabilities using a predefined projection matrix \( P \in [0,1]^{(G \times K)} \). The projection matrix transforms crop information into irrigation method probabilities based on regional farming practices. Let \( G = \{g_1, \dots, g_n\} \) be the set of \( n \) crop groups and \( \mathcal{Y} = \{1, \dots, K\} \) the set of \( K \) irrigation methods. The total area of crop group \( g \in G \) under irrigation type \( i \in \mathcal{Y} \) is denoted by \( A_{g,i} \). The probability \( P_{g,i} \) of irrigation type \( i \) for crop group \( g \) is given by:

\begin{equation}
 P_{g,i} =
 \begin{cases}
 \frac{A_{g,i}}{\sum_{j \in \mathcal{Y}} A_{g,j}}, & \text{if } \sum_{j \in \mathcal{Y}} A_{g,j} > 0 \\
 \frac{1}{K}, & \text{otherwise}
 \end{cases}
\end{equation}

where \( \sum_{i \in \mathcal{Y}} P_{g,i} = 1 \) for each \( g \in G\). The resulting \( P \in [0,1]^{n \times K} \) matrix is structured as:

\begin{equation}
P = \begin{bmatrix}
P_{1,1} & \hdots & P_{1,K} \\
\vdots & \vdots & \vdots \\
P_{n,1} & \hdots & P_{n,K} \\
\end{bmatrix}
\end{equation}

where each row represents a crop group's probability distribution across irrigation methods.
This predefined matrix offers three key advantages: $(i)$ it incorporates comprehensive historical knowledge about crop-irrigation relationships that cannot be learned from limited satellite imagery; $(ii)$ it provides reliable predictions even when irrigation infrastructure is not visible in the imagery; and $(iii)$ it enables easy updates of irrigation statistics without retraining the model.



    \item \textbf{Ensemble Module} combines data-driven predictions from Multi-Stream Module and knowledge-informed irrigation likelihoods using learnable weights $w$. Given two inputs (\(224\times224\times K\)), the module applies weighted summation to aggregate predictions, followed by softmax activation to normalize the final probability distribution. The weights are optimized jointly with the model parameters through end-to-end training. This enables the model to adaptively integrate spectral features with domain knowledge for more informed predictions.

\end{itemize}

\subsection{Loss Function}

We use a composite loss \( \mathcal{L} \) integrating cross-entropy loss (\(\mathcal{L}_c\)) \citep{jadon2020survey} for per-pixel classification and constrained Dice loss (\(\mathcal{L}_d\)) \citep{milletari2016v} to enforce spatial consistency:

\begin{equation}
\mathcal{L} = \alpha \mathcal{L}_c + (1-\alpha) \mathcal{L}_d,
\end{equation}

where \( \alpha \in [0,1] \) balances the two components.

For a given irrigation label \(Y\) and prediction \(\overline{Y}\), \textbf{cross-entropy loss} (\(\mathcal{L}_c\)) ensures pixel-wise class separation:

\begin{equation}
\mathcal{L}_c(Y, \overline{Y}) = -\frac{1}{H \times W} \sum_{k=1}^{K} \sum_{i,j} Y_{i,j,k} \log\left(\overline{Y}_{i,j,k}\right).
\end{equation}

\textbf{Land-masked dice loss} \(\mathcal{L}_d\) enhances segmentation quality by constraining predictions to agricultural land.

\begin{equation}
\mathcal{L}_d(Y, \overline{Y}) = 1 - \frac{1}{K} \sum_{k=1}^{K} \frac{2 \sum_{i,j} \overline{Y}_{i,j,k} \cdot Y_{i,j,k} \cdot L_{i,j}}{\sum_{i,j} L_{i,j} (Y_{i,j,k} + \overline{Y}_{i,j,k})},
\end{equation}
where \(L\) denotes the landmask. Land-masked dice loss enhances spatial coherence by considering neighboring pixels. In addition, it adjusts for class imbalance for underrepresented irrigation methods, such as drip irrigation, by normalizing over the sum of predictions and ground truth. Moreover, it constrains optimization to agricultural regions to prioritize learning irrigation patterns within farmland.

\begin{table*}[ht]
\scriptsize
\centering
\begin{tabular}{%
p{1.6cm}
p{.7cm}p{.5cm}p{.5cm}p{.7cm}|
p{.7cm}p{.5cm}p{.5cm}p{.7cm}|
p{.7cm}p{.5cm}p{.5cm}p{.7cm}|
p{.7cm}p{.5cm}p{.5cm}p{.7cm}
}
\toprule
\textbf{Model} & 
\multicolumn{4}{c}{\textbf{AZ}} &
\multicolumn{4}{c}{\textbf{UT}} &
\multicolumn{4}{c}{\textbf{WA}} &
\multicolumn{4}{c}{\textbf{CO}} \\
\cmidrule(lr){2-5}\cmidrule(lr){6-9}\cmidrule(lr){10-13}\cmidrule(lr){14-17}
& \textbf{MIoU} & \multicolumn{3}{c}{\textbf{Drip}} 
& \textbf{MIoU} & \multicolumn{3}{c}{\textbf{Drip}}
& \textbf{MIoU} & \multicolumn{3}{c}{\textbf{Drip}}
& \textbf{MIoU} & \multicolumn{3}{c}{\textbf{Drip}} \\
\cmidrule(lr){3-5}\cmidrule(lr){7-9}\cmidrule(lr){11-13}\cmidrule(lr){15-17}
& & P & R & IoU 
& & P & R & IoU
& & P & R & IoU
& & P & R & IoU \\
\midrule
\textbf{ResNet50}  
& 0.880 & 0.916 & 0.922 & 0.850  
& 0.542 & 0.625 & 0.117 & 0.110  
& 0.505 & 0.616 & 0.485 & 0.372  
& 0.740 & 0.866 & 0.661 & 0.600 \\

\textbf{LinkNet}     
& 0.878 & 0.914 & 0.925 & 0.851 
& 0.456 & 0.755 & 0.035 & 0.035  
& 0.539 & 0.612 & 0.583 & 0.425  
& 0.733 & 0.748 & 0.771 & 0.612 \\

\textbf{PAN}         
& 0.856 & 0.907 & 0.897 & 0.821 
& 0.561 & 0.590 & 0.394 & 0.309  
& 0.550 & 0.650 & 0.550 & 0.425  
& 0.699 & 0.743 & 0.617 & 0.508 \\

\textbf{FPN}         
& 0.836 & 0.897 & 0.871 & 0.792
& 0.556 & 0.625 & 0.311 & 0.262  
& 0.566 & 0.718 & 0.535 & 0.442  
& 0.722 & 0.782 & 0.684 & 0.575 \\

\textbf{DeepLabV3+}  
& 0.873 & 0.913 & 0.915 & 0.841  
& 0.564 & 0.448 & 0.316 & 0.228  
& 0.606 & 0.700 & 0.646 & 0.506  
& 0.747 & 0.796 & 0.712 & 0.602 \\

\textbf{ViT}         
& 0.751 & 0.873 & 0.831 & 0.742  
& 0.467 & 0.427 & 0.178 & 0.144  
& 0.478 & 0.600 & 0.398 & 0.314  
& 0.626 & 0.782 & 0.491 & 0.432 \\

\textbf{FarSeg}      
& 0.878 & 0.930 & 0.903 & 0.846
& 0.575 & 0.575 & 0.374 & 0.293  
& 0.554 & 0.682 & 0.556 & 0.441  
& 0.769 & 0.850 & 0.742 & 0.656 \\

\textbf{SegFormer}   
& 0.867 & 0.918 & 0.901 & 0.833
& 0.577 & 0.580 & 0.425 & 0.325  
& 0.558 & 0.720 & 0.528 & 0.438  
& 0.727 & 0.833 & 0.651 & 0.576 \\

\textbf{Swin}        
& 0.896 & 0.942 & 0.913 & 0.865  
& 0.640 & 0.472 & 0.584 & 0.353  
& 0.645 & 0.752 & 0.663 & 0.544  
& 0.786 & 0.810 & 0.797 & 0.671 \\

\midrule
\textbf{KIIM}     
& \textbf{0.988} & \textbf{0.988} & \textbf{0.993} & \textbf{0.982}
& \textbf{0.791} & \textbf{0.664} & \textbf{0.873} & \textbf{0.605}
& \textbf{0.770} & \textbf{0.796} & \textbf{0.776} & \textbf{0.647}
& \textbf{0.931} & \textbf{0.820} & \textbf{0.967} & \textbf{0.798} \\
\bottomrule
\end{tabular}
\caption{Macro-IoU (averaged over Flood, Sprinkler, Drip, and non-irrigated land) and Drip-only performance (without Dice) for each model in AZ, UT, WA, and CO.}
\label{tab:tab:state-wise-performance}
\end{table*}
\section{Experimentation and Results}

\subsection{Dataset}
Our study integrates multi-source geospatial datasets to construct an irrigation and crop mapping framework across five U.S. states, using the Utah Water-Related Land Use (WRLU) Dataset (2023) for Utah\footnote{\url{https://dwre-utahdnr.opendata.arcgis.com/pages/wrlu-data}}, the USGS Verified Irrigated Agricultural Lands Dataset (2002–2017) for Arizona and Florida\footnote{\url{https://catalog.data.gov/dataset/verified-irrigated-agricultural-lands-for-the-united-states-200217}}, the Washington State Department of Agriculture Agricultural Land Use dataset for Washington\footnote{\url{https://agr.wa.gov/departments/land-and-water/natural-resources/agricultural-land-use}}, and the Colorado Division of Water Resources GIS dataset for Colorado\footnote{\url{https://dwr.colorado.gov/services/data-information/gis}}. In addition, we map various irrigation practices to three primary methods (drip, sprinkler, and flood), as the original datasets contained multiple irrigation subtypes that were unified for consistency in this study. Crop data are derived from these sources, consolidating 143 distinct crop types into 20 standardized categories (details in Appendix). In addition, we collect Landsat-8 satellite imagery to generate irrigation masks with a spatial resolution of 30 meters.

Following standard remote sensing procedures, we segment satellite images into non-overlapping patches of size 224×224 pixels. The crop mask is created by assigning each pixel to a crop group based on the mapped crop type, while the land mask is derived by categorizing pixels into agricultural and nonagricultural lands. The final dataset comprises 36738 image patches, including Arizona (7154 patches), Utah (6062 patches), Washington (3557 patches), Florida (1230 patches) and Colorado (18735 patches). Among the state datasets, Utah and Florida have the lowest percentage (1.8\% and 0.08\%) of patches with drip irrigation. We discuss further dataset collections, dataset details, preprocessing steps, and projection matrix formulation in Appendix.
\subsection{Evaluation Metrics}
We evaluate the irrigation mapping task using four standard segmentation metrics: Intersection over Union (IoU), Precision (P), Recall (R), and Dice Score (D). Let \( Y \) and \( \overline{Y} \) be the ground truth and predicted masks, respectively, for an image of size \( H \times W \), where each pixel \((i,j)\) is assigned a class \( k \in \mathcal{Y} \). The ground truth and predicted pixel sets for class \( k \) are defined as:
\begin{equation}
    T_k = \{(i, j) \mid Y_{i,j} = k\}, \quad 
    M_k = \{(i, j) \mid \overline{Y}_{i,j} = k\}.
\end{equation}

The evaluation metrics precision, recall, Dice, and IoU are computed as:
\begin{align}
    \text{P}_k = \frac{|M_k \cap T_k|}{|M_k|}, \quad \quad \quad
    \text{R}_k &= \frac{|M_k \cap T_k|}{|T_k|},\\
    \text{D}_k = \frac{2 \times P_k \times R_k}{P_k+R_k},
    \quad \quad
    \text{IoU}_k &= \frac{|M_k \cap T_k|}{|M_k \cup T_k|},
\end{align}
Precision measures the proportion of correctly predicted irrigated pixels among all predicted as class \( k \), while recall quantifies the fraction of correctly identified irrigated pixels out of all actual class \( k \) pixels. On the contrary, Dice Score computes the harmonic mean of precision and recall. IoU is defined as the ratio of intersection to union, provides a more balanced spatial evaluation by penalizing both over-segmentation (false positives) and under-segmentation (false negatives).


\subsection{Experimentation Setting}
We split each state's dataset into 85\% training and 15\% testing, except for Florida, where we used a 50\%-50\% split due to limited drip irrigation samples. Model implementation was conducted using PyTorch and executed on NVIDIA A40 GPU. We performed 5-fold cross-validation and optimized hyperparameters through grid search over learning rates \( \{1e\text{-}4, 2e\text{-}4, 5e\text{-}4\} \), batch sizes \( \{16, 32, 64\} \), and loss weight \( \alpha \) values \( \{0, 0.4, 0.5, 0.6, 1\} \). The optimal configuration was selected based on IoU performance on the validation set.


\subsection{Effectiveness of KIIM Model}

We first evaluate KIIM’s ability to classify irrigation methods, particularly drip irrigation, and compare it against nine state-of-the-art segmentation models, including transformer-based architectures (Swin, ViT, SegFormer) and the remote sensing-specific model FarSeg.
We train each model on state-specific irrigation datasets and evaluate on the corresponding test dataset across Arizona (AZ), Utah (UT), Washington (WA) and Colorado (CO). In Table \ref{tab:tab:state-wise-performance}, we show the performance of our KIIM model and state-of-the-art models. Across all states, state-of-the-art models struggle to identify drip-irrigated lands due to extreme class imbalance. The best baseline (Swin) achieves only 0.353 and 0.544 IoU in Utah and Washington which highlights the challenges of identifying drip irrigation. Due to its sparse presence and high spectral similarity with other irrigated lands, state-of-the-art models fail to identify drip irrigated lands. In contrast, the higher prevalence of drip irrigation in Arizona makes it easier for models to learn its spatial patterns and distinguish it from other irrigation methods.

\begin{figure}[h]
    \centering
    \includegraphics[width=0.99\linewidth]{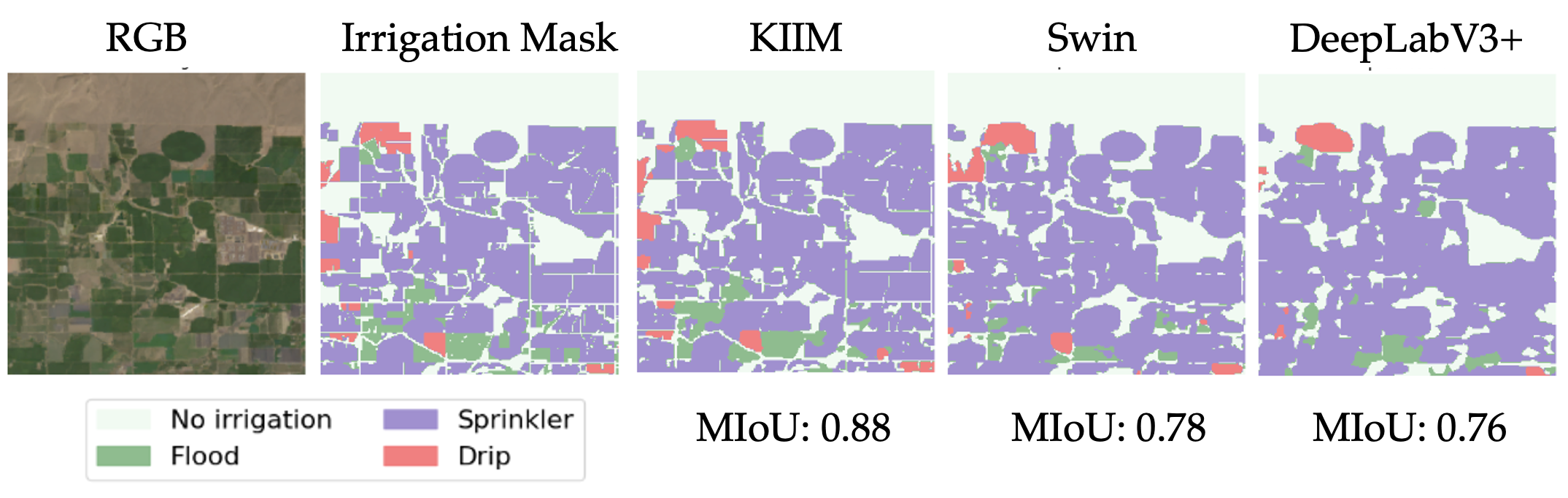}
   \caption{Visual comparison of our model (KIIM) against top-performing baselines (Swin and DeepLabV3+). KIIM accurately segments farmland and correctly classifies irrigation methods, whereas baseline models struggle to delineate agricultural farmland and misclassify irrigation methods. The higher MIoU scores underscores KIIM model's effectiveness in detecting and identifying different irrigation methods.}
    \label{fig:example}
\end{figure}

We notice that the KIIM model effectively captures spatial dependencies and distinguishes underrepresented irrigation methods, which leads to significantly improved segmentation accuracy. Specifically, KIIM outperforms the best baseline (Swin) in macro-IoU by 10.3\% in AZ (0.988 vs. 0.896), 19.6\% in CO (0.931 vs. 0.778), 22.9\% in UT (0.791 vs. 0.644), and 19.4\% in WA (0.770 vs. 0.645). Moreover, KIIM achieves a 13.5\% improvement in AZ (0.982 vs. 0.865), 18.9\% in CO (0.798 vs. 0.671), 71.4\% in UT (0.605 vs. 0.353), and 19.0\% in WA (0.647 vs. 0.544). The most significant gain is observed in Utah,  where baseline models struggle due to severe class imbalance, but KIIM improves drip IoU from 0.353 (Swin) to 0.605 (a 71.4\% improvement). In Figure \ref{fig:example}, we present an example of KIIM model’s predictions, demonstrating the model's ability to accurately classify irrigation methods These results highlight KIIM’s robustness in handling extreme class imbalances and its effectiveness in different geographical regions. 

\subsection{Transfer Learning for Cross-State Irrigation Mapping}
Mapping irrigation methods across different states is challenging due to limited labeled data, extreme class imbalance, and substantial regional variations in irrigation practices. Therefore, training separate state-specific models is often impractical, as some irrigation methods (e.g., drip irrigation) are severely underrepresented in many states. Moreover, in some cases, sufficient training data is entirely unavailable. For example, Florida has only 1230 training samples, with only 11 samples containing drip irrigation, which makes it impossible to train and test a reliable state-specific model. To overcome this, we leverage transfer learning that enables a model to learn common irrigation features from a diverse multi-state dataset and adapt them to specific states with minimal labeled data.


In this work, we create a multi-state training dataset which maintains enough drip irrigated samples in the training data. Following \cite{buda2018systematic}, we construct the multi-state training dataset (\(8880\) samples where \(4440\) samples have drip irrigation) where the total sample size is twice the number of drip-irrigated samples, maintaining an imbalance ratio of $(2:1)$. The training samples were selected from the Arizona, Utah, Colorado, and Washington training data. 


To train our model, we follow a two-phase learning approach (training and state-adaptive fine-tuning). In the \textbf{training step}, we train KIIM on the multi-state dataset to learn universal irrigation patterns (e.g., circular sprinkler layouts) that are consistent across states. In the \textbf{state-adaptive fine-tuning} step, we initialize state-specific models with pretrained weights (from the training step) and fine-tune them using state-specific data to adapt to regional farm sizes, irrigation preferences, and class distributions while keeping the architecture unchanged. This hierarchical learning strategy enables KIIM to generalize across states while capturing local irrigation nuances and improving segmentation performance with minimal state-level labeled data.

\begin{table}[ht]
\scriptsize
\centering
\small
\begin{tabular}{p{.3cm}p{0.95cm}|p{.6cm}p{.7cm}|p{.6cm}p{.7cm}|p{.7cm}|p{.6cm}p{.7cm}}
\toprule
\textbf{State} & \textbf{Model} 
& \multicolumn{2}{c|}{\textbf{Flood}} 
& \multicolumn{2}{c|}{\textbf{Sprinkler}} 
& \multicolumn{2}{c}{\textbf{Drip}} \\
\cmidrule(lr){3-4} \cmidrule(lr){5-6} \cmidrule(lr){7-8}
& & \textbf{Dice} & \textbf{IoU} 
& \textbf{Dice} & \textbf{IoU} 
& \textbf{Dice} & \textbf{IoU} \\
\midrule

\multirow{3}{*}{\textbf{AZ}} 
& KIIM         & 0.991 & 0.983 & 0.995 & 0.989 & 0.991 & 0.982 \\
& w/o FT   & 0.982 & 0.964 & 0.993 & 0.986 & 0.985 & 0.971 \\
& w FT      & 0.986 & 0.973 & 0.994 & 0.988 & 0.988 & 0.976 \\
\midrule

\multirow{3}{*}{\textbf{UT}} 
& KIIM         & 0.876 & 0.780 & 0.893 & 0.806 & 0.754 & 0.605 \\
& w/o FT   & 0.722 & 0.565 & 0.766 & 0.620 & 0.613 & 0.442 \\
& w FT      & 0.884 & 0.792 & 0.900 & 0.819 & 0.835 & 0.717 \\
\midrule

\multirow{3}{*}{\textbf{WA}} 
& KIIM         & 0.759 & 0.611 & 0.926 & 0.863 & 0.786 & 0.647 \\
& w/o FT   & 0.702 & 0.541 & 0.928 & 0.866 & 0.780 & 0.640 \\
& w FT      & 0.820 & 0.695 & 0.945 & 0.896 & 0.838 & 0.722 \\
\midrule

\multirow{3}{*}{\textbf{CO}} 
& KIIM         & 0.983 & 0.966 & 0.981 & 0.963 & 0.887 & 0.798 \\
& w/o FT   & 0.950 & 0.904 & 0.932 & 0.873 & 0.852 & 0.743 \\
& w FT      & 0.984 & 0.969 & 0.980 & 0.961 & 0.937 & 0.882 \\
\bottomrule
\end{tabular}
\caption{Performance of KIIM model for state-wise training (denoted as KIIM), cross-state training without state-adaptive fine-tuning (denoted as w/o FT), and cross-state training with state-adaptive fine-tuning (denoted as w FT) for Flood, Sprinkler, and Drip (Dice and IoU).}
\label{tab:kiim-finetuning}
\end{table}
\textbf{Effectiveness of cross-state transfer learning:} To assess the effectiveness of cross-state knowledge transfer, we compare two training strategies for KIIM: $(i)$ a two-step transfer learning approach, where the model is first trained on a balanced multi-state dataset and then fine-tuned on state-specific data, and $(ii)$ state-specific training only (KIIM). Additionally, we evaluate the model without the fine-tuning setting (zero-shot), where the model is pretrained on the multi-state dataset but not fine-tuned on the target state. The results in Table \ref{tab:kiim-finetuning} demonstrate that transfer learning effectively enhances both majority and minority class segmentation. Performance on the majority class (sprinkler) remains consistent across approaches (AZ: 0.989, UT: 0.819, WA: 0.863, and CO: 0.961 IoU). However, minority class performance improves significantly (particularly for drip irrigation), where IoU increases from 0.605 to 0.717 in Utah (18.5\%), and from 0.647 to 0.722 in Washington (11.5\%). Similarly, in Washington, flood IoU improves from 0.611 to 0.695 (13.7\%). This indicates that transfer learning effectively captures underrepresented irrigation patterns. It is noteworthy that the model without state-specific adaptations achieves similar performance as compared to the baseline (state-wise training only) for the majority class (e.g., sprinkler). These results validate that cross-state transfer learning retains general irrigation knowledge while adapting to state-specific variations.

\begin{figure}[!htb]
    \centering
    \includegraphics[width=\linewidth]{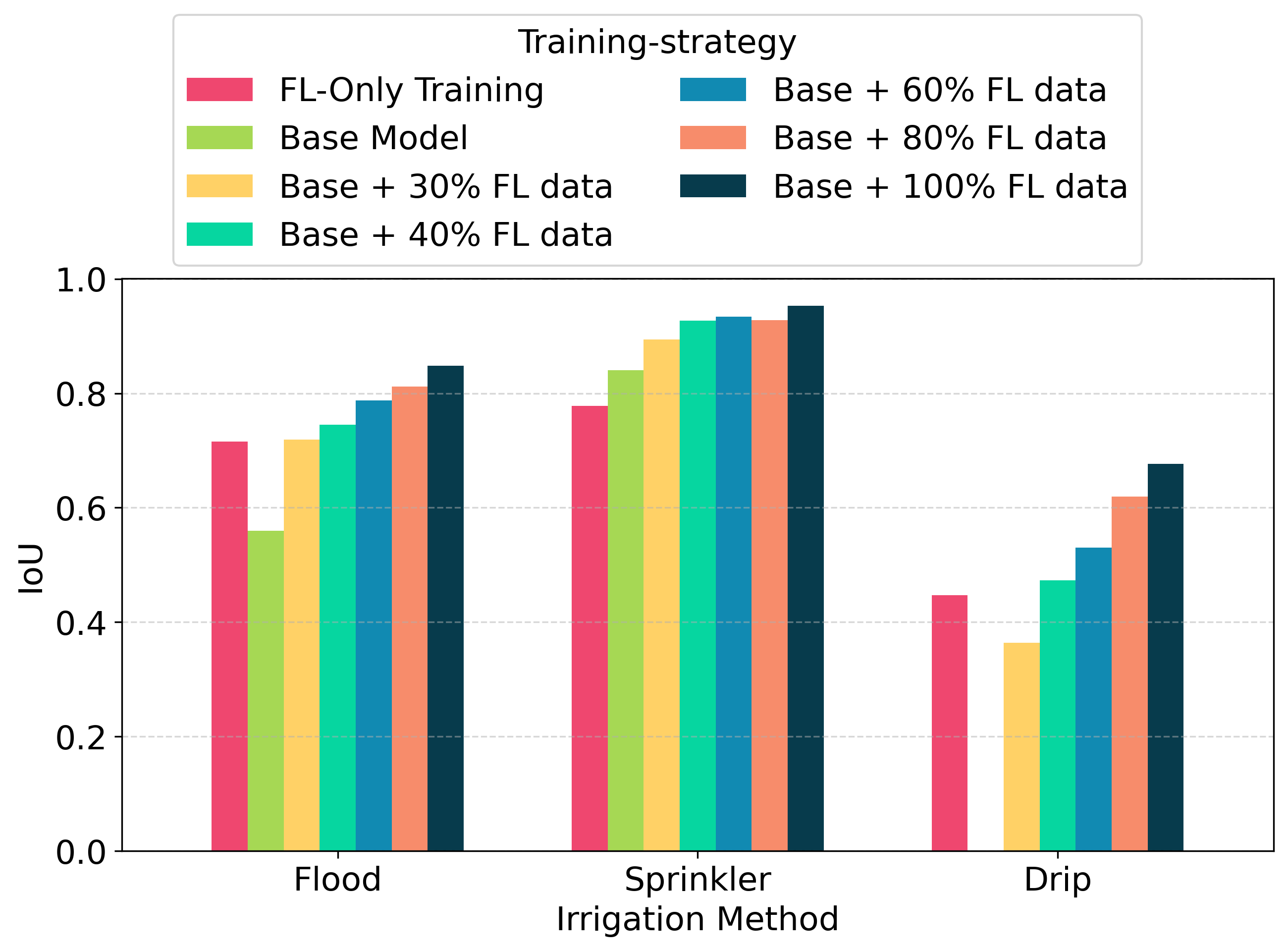}
    \caption{KIIM model performance for the state of Florida for different training approaches. `FL-Only Training' indicates that KIIM is trained solely on the Florida training dataset, whereas `Base + X\% data' indicates that KIIM is trained on a combined dataset (without Florida) and fine-tuned on X\% of the Florida training dataset. Notably, for drip irrigation, KIIM attains a 0.678 IoU score with cross-state transfer learning which indicates 51\% improvement over state-wise training.}
    \label{fig:florida-results}
\end{figure}

\textbf{Generalization irrigation mapping across states:} The Florida data set is very sparse and has severe class imbalance which makes state-only training ineffective for irrigation mapping. To address this, we implement cross-state transfer learning, where KIIM is pretrained on a multi-state irrigation dataset and fine-tuned with incremental portions of Florida training data (30\%, 40\%, 60\%, 80\%, and 100\%). This approach allows the model to leverage universal irrigation features learned from diverse states (e.g., AZ, WA, CO, and UT) and improve the mapping task even with small amounts of labeled data. Figure \ref{fig:florida-results} shows that our model achieves IoU of 0.56 and 0.86 for flood and sprinkler irrigation without seeing any data from Florida. This demonstrates that irrigation structures exhibit transferable patterns across regions. Moreover, tuning the model with Florida data, the performance improves significantly. For drip irrigation (minority class), the model achieves IoU 0.678 when tuned on all the training data. This underscores the necessity of state-specific fine-tuning to capture local irrigation practices. In comparison to the baseline (state-wise training), the IoU improves from 0.447 to 0.678 (51\%) for drip irrigation. Notably, even with 40\% fine-tuning, performance is comparable to full-state training. This suggests our model can be effectively used for irrigation mapping when very limited data are available. These findings show that cross-state learning enables the model to generalize irrigation patterns across regions with minimal labeled data.

\section{Ablation Study}
To evaluate the individual contribution of each architectural component and validate our design choices, we conduct an ablation study by systematically removing different modules from our KIIM model. Our results demonstrate that each module plays an important role in irrigation type mapping. From Table \ref{tab:ablation-study}, we show that the model, without any specialized modules, achieves a macro IoU score of 0.712, while our complete architecture incorporating all components reaches 0.883. Similarly, the exclusion of any module reduces the macro IoU score. This highlights the importance of land mask information, crop information, land-masked Dice loss, and multistream module for identifying irrigation patterns.
\begin{table}[h]
\scriptsize
    \centering
    \begin{tabular}{cccccc}
        \toprule
        AM & PM & LDL & MSM & Dice & IoU \\
        \midrule
        \ding{51} & \ding{51} & \ding{51} & cross & \textbf{0.937} & \textbf{0.883} \\
        \ding{51} & \ding{51} & \ding{51} & self & 0.931 & 0.873 \\
        
        \ding{55} & \ding{51} & \ding{51} & cross & 0.850 & 0.747 \\
        \ding{55} & \ding{55} & \ding{51} & cross & 0.842 & 0.736 \\
        \ding{55} & \ding{51} & \ding{55} & cross & 0.853 & 0.753 \\
        \ding{55} & \ding{55} & \ding{55} & cross & 0.844 & 0.739 \\
        \ding{55} & \ding{55} & \ding{55} & \ding{55} & 0.826 & 0.712 \\ 
        \bottomrule
    \end{tabular}
    \caption{Performance of KIIM on validation data while varying different modules. AM indicates attention module, PM denotes projection module, LDL indicates land-masked dice loss, and MSM indicates multi-stream module technique. Checkmarks (\ding{51}) indicate inclusion of the respective module, while crosses (\ding{55}) indicate exclusion. Note that we report macro IoU, and macro Dice scores in the Table.}
    \label{tab:ablation-study}
\end{table}
\noindent
\section{Discussion}
While accurate irrigation mapping is crucial for identifying current practices and guiding sustainable upgrades, existing approaches rely on manual surveys (e.g., USGS data) or lack generalization across different regions and irrigation methods. To address this, we propose the Knowledge-Informed Irrigation Mapping (KIIM) model, a multi-stream framework that integrates RGB and agriculture-specific indices through a bidirectional attention module for enhanced feature fusion across different modalities of input streams. Additionally, KIIM incorporates land-use and crop data, enabling the model to focus on agriculture-specific pixels and leverage historical crop-irrigation relationships for improved classification. Our findings highlight the effectiveness of the proposed approach in improving irrigation mapping across multiple states, particularly in challenging cases like drip irrigation. The substantial performance gains over the baseline demonstrate the model’s ability to capture complex spatial patterns and stream-specific relationships. {Therefore, KIIM enables timely identification of irrigation, which directly contributes to SDG 2 by supporting sustainable food production systems and resilient agricultural practices in water-stressed regions.}

However, part of our model's performance relies on historical crop-irrigation data, which may change over time. Also, crop-type (and land-use) masks may be erroneous or outdated in certain regions. Future work could explore 
more representative region-specific datasets and extending this framework to more diverse agricultural landscapes and refining it for even greater adaptability across varying irrigation practices.

\section*{Acknowledgments}
{This material is based upon work supported by the AI Research Institutes program supported by NSF and USDA-NIFA under the AI Institute: Agricultural AI for Transforming Workforce and Decision Support (AgAID) award No.~2021-67021-35344. This work was
partially supported by University of Virginia Strategic Investment Fund award number SIF160.}

\bibliographystyle{named}
\bibliography{ijcai25}

\appendix
\clearpage
\section{Background}
\subsection{Agricultural Related Spectral Bands}
\label{agbands}
In this work, we use three agriculture-related spectral bands: ($i$) normalized difference tillage index (NDTI), ($ii$) normalized difference water index (NDWI), and ($iii$) normalized difference vegetation index (NDVI).

\textbf{NDTI} quantifies the agricultural preparation of soil by measuring the difference between two short wave infrared (SWIR) satellite bands.

\begin{equation}
 \mathrm{NDTI} = \frac{\mathrm{SWIR1} - \mathrm{SWIR2}}{\mathrm{SWIR1} + \mathrm{SWIR2}}
\end{equation}
This is used for the assessment of crop residue cover and tillage practices and differentiation between crop and non-crop areas \cite{zheng2014remote} \cite{van2015using}.

\textbf{NDWI} is used to detect water bodies in satellite or aerospace images. It is based on the difference in light absorption in the near-infrared (NIR) and visible green ranges of the electromagnetic spectrum.
\begin{equation}
 \mathrm{NDWI} = \frac{\mathrm{NIR} - \mathrm{SWIR1}}{\mathrm{NIR} + \mathrm{SWIR1}}
\end{equation}
Generally, NDWI is used to estimate vegetation water, wetland delineation, and drought monitoring \cite{gao1996ndwi}\cite{gu2007five} \cite{mcfeeters1996use}.

\textbf{NDVI} is widely used for quantifying the health and density of vegetation using sensor data. It is calculated as follows:

\begin{equation}
 \mathrm{NDVI} = \frac{\mathrm{NIR} - \mathrm{Red}}{\mathrm{NIR} + \mathrm{Red}}
\end{equation}
These spectral bands are widely used for crop health monitoring, yield prediction, drought assessment, and land cover classification. \cite{pettorelli2005using} \cite{kogan1995application} \cite{tucker1985satellite}.
\section{Dataset} 
\subsection{Irrigation Data Collection}
In this study, we utilize an irrigation mapping dataset from five states: $(i)$ Utah, $(ii)$ Arizona, $(iii)$ Washington, $(iv)$ Colorado, and $(v)$ Florida. The dataset was collected from four sources as follows:

\textbf{Utah Water Related Landuse Dataset:}
The Water-Related Land Use (WRLU)\footnote{https://dwre-utahdnr.opendata.arcgis.com/pages/wrlu-data} dataset from 2023 provides detailed vector polygons of irrigated fields in Utah, including irrigation methods, crop types, water sources, and acreage information. In 2023, the irrigated agricultural lands (1,735,422 acres) were distributed across three irrigation methods: drip (0.08\%), sprinkler (34\%), and flood (31\%). Due to the limited representation of drip irrigation in 2023, we incorporated drip-irrigated lands data from 2017 to 2022.

\textbf{Multi-state Dataset (AZ and FL):}
For Arizona and Florida, we utilized the USGS Verified Irrigated Agricultural Lands dataset (2002-2017)\footnote{https://catalog.data.gov/dataset/verified-irrigated-agricultural-lands-for-the-united-states-200217}, a geographic information system (GIS) geodatabase developed collaboratively by USGS and the University of Wisconsin. We use Washington State Department of Agriculture Agricultural Land Use dataset for Washington, and the Colorado Division of Water Resources GIS dataset for Colorado. This data for each state has been collected through multiple years. Specifically,
\begin{itemize}
    \item \textbf{Arizona}: 2013-2017 data, covering 12 irrigation practices.
    \item \textbf{Washington}: 2023 data, containing 9 irrigation methods.
    \item \textbf{Colorado}: 2018-2020 data, covering 4 irrigation practices.
    \item \textbf{Florida}: 2014-2017 data, covering 5 irrigation practices.
\end{itemize}
To standardize the analysis across states, we mapped various irrigation practices to three primary methods: drip, sprinkler, and flood irrigation. The distribution of irrigation methods varies significantly across the states. In Arizona's 566,340 acres of agricultural lands, 8\% use drip irrigation, 46\% flood irrigation, and 20\% sprinkler irrigation. Washington's 7,748,932 acres show a different pattern with 2\% drip irrigation, 2\% flood irrigation, and 21\% sprinkler irrigation. Florida's 548,010 acres are predominantly flood-irrigated (25\%) followed by sprinkler (22\%), and drip (0.01\%) using drip irrigation. Similarly, Colorado's 2,560,487 acres are predominatly irrigated using flood irrigation.

\subsection{Crop Data Collection} \label{sec:crop-data}
A total of 143 distinct crop types were identified across the five states from the WRLU and USGS Verified Irrigated Agricultural Lands datasets. To standardize the analysis, we consolidated these crops into 20 categorical groups based on classifications from Leff et al. \cite{leff2004geographic} and the IR4-Project\footnote{https://www.ir4project.org/fc/crop-grouping/crop-group-tables/} of U.S. Department of Agriculture (USDA). The groups are: Alfalfa, Cereals, Cover Crop, Fibres, Fruits, Grass, Green House, Herb Group, Horticulture, Nursery, Nuts, Oil-bearing crops, Orchard, Pulses, Roots and Tubers, Shrub Plants, Sugar Crops, Vegetables, Vineyard, and an additional category for unspecified crops. For instance, the cereal group includes barley, corn, wheat, and sorghum, while the fruit group encompasses apples, berries, citrus, and melons. The complete mapping of individual crops to their respective groups is provided in the Appendix Table \ref{tab:crop_mapping}.
\subsection{Agricultural Land Data Colection}
The study area encompasses various land use categories including irrigation, dry agriculture, idle, riparian, sub-irrigation, urban, urban grass, water, and wet flats. From these categories, we focused specifically on irrigated lands, agricultural lands, and urban areas, as these represent the primary zones requiring active irrigation management.

\subsection{Satellite Data Collection}
We collected Landsat-8 satellite imagery from USGS Earth Explorer\footnote{https://earthexplorer.usgs.gov/} for the five states (Utah, Arizona, Washington, Colorado, and New Mexico) corresponding to their respective study periods. Landsat-8 carries two instruments: the Operational Land Imager (OLI) and the Thermal Infrared Sensor (TIRS), providing nine spectral bands at 30m spatial resolution and two thermal bands at 100m. The satellite has a 16-day repeat cycle, and we focused on data acquisition during the irrigation season (March to September). Images with significant (more than 5\%) cloud cover or poor quality, identified through the Quality Assessment (QA) band, were excluded from the analysis.

\subsection{Dataset Preparation}
For the creation of irrigation masks, we preprocessed each Landsat scene by projecting it to the WGS 84 coordinate system (EPSG:4326). The scenes were then segmented into non-overlapping patches of 224×224 pixels, where each patch covers approximately 45 square kilometers at a 30-meter resolution. We performed pixel-wise mapping of each patch to corresponding irrigation types based on the available ground truth data.
To ensure data quality, we implemented a multi-stage filtering process:

\begin{itemize}
    \item Discarded patches where more than 95\% of pixels were labeled as non-irrigated, ensuring each selected patch contains meaningful irrigation information.
    \item Removed patches affected by snow, clouds, or cloud shadows through manual inspection of the imagery.
    \item Excluded patches containing incomplete or ambiguous irrigation labels to maintain data integrity.
\end{itemize}
 We further create a projection matrix for each studied state. For each state, we consider all available crop groups and associated lands for each irrigation method to find the projection matrix.


\begin{table*}[h!]
    \centering
    \caption{Mapping of Original Irrigation Labels to Standardized Categories.}
    \begin{tabular}{|p{15cm}|c|}
    \hline
    \textbf{Original Label} & \textbf{Mapped Label} \\ \hline
    Traveler Sprinkler, Center Pivot - Tow, Solid State Sprinkler, Overhead, Traveling Gun, Pivot, Lateral Sprinkler, 
    Other Sprinkler, Big Gun, Wheel Line, Big Gun/Sprinkler, Sprinkler/Wheel Line, Center Pivot, Micro-Sprinkler, 
    Micro-Bubbler, Sprinkler \& Bubbler, Lateral, Side Roll, Center Pivot/Sprinkler, Center Pivot/Wheel Line, 
    Big Gun/Wheel Line, Big Gun/Sprinkler/Wheel Line & Sprinkler \\ \hline
    Drip Microirrigation, Micro-Drip & Drip \\ \hline
    Furrow, Grated$\_$Pipe, Improved Flood, Rill, Hand/Rill, None/Rill, Gated$\_$pipe & Flood \\ \hline
    Not Specified, Micro, Research, Uncertain, Drip/None, Big Gun/Drip, Drip/Big Gun, Drip/Rill/Sprinkler, 
    Rill/Sprinkler, Drip/Micro-Sprinkler, Drip/Wheel Line, Center Pivot/Rill, Rill/Wheel Line, Drip/Rill, Center Pivot/None, 
    Center Pivot/Rill/Wheel Line, Center Pivot/Sprinkler/Wheel Line, Center Pivot/Rill/Sprinkler, 
    Rill/Sprinkler/Wheel Line, Center Pivot/Drip, Hand/Sprinkler, Drip/Sprinkler, Sub-irrigated, Dry Crop, 
    Sprinkler And Drip, Center Pivot/Drip/Sprinkler, Unknown, Non$\_$irrigated & Removed \\ \hline
    \end{tabular}
    \label{tab:irrigation_label_mapping}
\end{table*}

\begin{table*}
\centering
\caption{Mapping of Individual Crops to Crop Groups. The `UNK' crop group indicates the crops can not be specified in any crop groups.}
\label{tab:crop_mapping}
\resizebox{\textwidth}{!}{
\begin{tabular}{ll}
\hline
\textbf{Crop Group} & \textbf{Individual Crops} \\
\hline
Alfalfa & Alfalfa, Alfalfa/Barley Mix, Alfalfa/Grass, New Alfalfa \\
\hline
Cereals & Barley, Barley/Wheat, Cereal Grain, Corn, Durum Wheat, Grain/Seeds unspecified, Oats, Rye, \\
& Sorghum, Speltz, Spring Wheat, Triticale, Wheat, Winter Wheat, Corn Grain, Corn Silage, \\
& Small Grains, Sorghum Grain, Spring Grain, Sweet Corn, Wheat Fall, Wheat Spring, Field Corn, \\
& Double crop barley/corn, Double crop winter wheat/corn \\
\hline
Cover Crop & Cover Crop, Green Manure, Field Crops, Other Field Crops \\
\hline
Fibres & Cotton \\
\hline
Fruits & Apples, Apricots, Berries, Berry, Cherries, Citrus, Dates, Fruit Trees, Grapes, Melon, \\
& Oranges, Peaches, Pomegranate, Citrus Groves, Fruit \\
\hline
Grass & Bermuda Grass, Grass, Grass Hay, Hay/Silage, Idle Pasture, Other Hay/Non Alfalfa, Pasture, \\
& Pecan/Grass, Sod, Turfgrass, Turfgrass Ag, Turfgrass Urban, Grass Pasture, Bluegrass, \\
& Sod Farm, Grass/Hay/Pasture, Hay, Improved Pasture - Irrigated, Rye Grass, Grassland/Pasture, \\
& Irrigated turf \\
\hline
Green House & Greenhouse \\
\hline
Herb Group & Flowers, Herb \\
\hline
Horticulture & Horticulture \\
\hline
Nursery & Nurseries, Nursery, Nursery Trees, Tree Nurseries, Tree Nursery \\
\hline
Nuts & Almond, Pecans, Pistachios, Walnuts \\
\hline
Oil-bearing crops & Canola, Flaxseed, Jojoba, Mustard, OilSeed, Olives, Safflower, Soybeans \\
\hline
Orchard & Orchard, Orchard unspecified, Orchard With Cover, Orchard Wo Cover \\
\hline
Pulses & Beans, Dry Beans, Garbanzo, Seed, Peanuts, Seeds \\
\hline
Roots and Tubers & Potato, Potatoes \\
\hline
Shrub Plants & Guayule, Shrubland \\
\hline
Sugar Crops & Sugar Beets, Sugarbeets, Sunflower, Sugar Cane, Sugar cane \\
\hline
UNK & Commercial Tree, Fallow, Fallow/Idle, Field Crop unspecified, Idle, Not Specified, Other, \\
& Sudan, Transition, Trees, Urban, Ornamentals, Research Facility, Research land, \\
& Miscellaneous vegetables and fruits, Other tree crops \\
\hline
Vegetables & Flower Bulb, Lettuce, Onion, Pumpkins, Squash, Vegetable, Vegetables, Watermelons, \\
& Eggplant, Fall Vegetables, Spring Vegetables, Vegetables Double Crop, Cabbage, Onions, Peppers \\
\hline
Vineyard & Vineyard \\
\hline
\end{tabular}
}
\end{table*}

\end{document}